%% file: root.tex
\documentclass[letterpaper, 10 pt, conference]{ieeeconf}  

\usepackage{graphicx}
\usepackage{comment}
\usepackage[table]{xcolor}  
\usepackage{colortbl}       
\usepackage{amsmath}        
\usepackage{pifont}         
\usepackage{booktabs}       
\usepackage{wrapfig}        
\usepackage{multirow}       
\usepackage{array}          
\usepackage{hyperref}       

\usepackage{cite}  
\usepackage{hyperref}
\hypersetup{hypertex=true,
            colorlinks=true,
            linkcolor=black,
            anchorcolor=black,
            citecolor=black}

\definecolor{zyx_color}{rgb}{0.5,0.2,0.8}

\definecolor{ltb_color}{rgb}{0.2,0.8,0.8}

\definecolor{need_check}{rgb}{1.0,0,0}            
\newcommand{\needCheck}[1]{\textcolor{need_check}{#1}} 

\newcommand{\toolName}{\textit{DynamicPose}} 

\IEEEoverridecommandlockouts                              

\title{\LARGE \bf

\toolName{}: Real-time and Robust 6D Object Pose Tracking for Fast-Moving Cameras and Objects}

\author{
  Tingbang Liang\textsuperscript{1,2,*},
  Yixin Zeng\textsuperscript{1,*},
  JiaTong Xie\textsuperscript{1},
  and Boyu Zhou\textsuperscript{3,4,\textdagger}
  \thanks{\textbf{* Equal contribution. Listing order is random.}}%
  \thanks{\textdagger\ \textbf{Corresponding Author.}}%
  \thanks{\textsuperscript{1}School of Artificial Intelligence, Sun Yat-Sen University, Zhuhai, China.}%
  \thanks{\textsuperscript{2}School of Mechanical Engineering, Xi'an Jiaotong University, Xi'an, China.}%
  \thanks{\textsuperscript{3}Department of Mechanical and Energy Engineering, Southern University of Science and Technology, Shenzhen, China.}%
  \thanks{\textsuperscript{4}Differential Robotics.}%
  \thanks{Project supported by the Young Scientists Fund of the National Natural Science Foundation of China (Grant No. 62403502).}%
}

\begin{document}

\maketitle

\thispagestyle{empty}
\pagestyle{empty}

\input{sec/0_abstract}

\input{sec/1_introduction}
\input{sec/2_related_work}

\input{sec/3_problem_formulation}
\input{sec/4_method}
\input{sec/5_experiments}
\input{sec/6_discussion_and_future_work}




\bibliographystyle{IEEEtran}
\bibliography{IEEEtranBST/OurRef}

\end{document}

%% file: sec/0_abstract.tex
\begin{abstract}
We present \toolName{}, a retraining-free 6D pose tracking framework that improves tracking robustness in fast-moving camera and object scenarios.
Previous work is mainly applicable to static or quasi-static scenes, and its performance significantly deteriorates when both the object and the camera move rapidly.
To overcome these challenges, we propose three synergistic components:
(1) A visual-inertial odometry compensates for the shift in the Region of Interest (ROI) caused by camera motion;
(2) A depth-informed 2D tracker corrects ROI deviations caused by large object translation;
(3) A VIO-guided Kalman filter predicts object rotation, generates multiple candidate poses, and then obtains the final pose by hierarchical refinement.
The 6D pose tracking results guide subsequent 2D tracking and Kalman filter updates, forming a closed-loop system that ensures accurate pose initialization and precise pose tracking. 
Simulation and real-world experiments demonstrate the effectiveness of our method, achieving real-time and robust 6D pose tracking 
for fast-moving cameras and objects.
\end{abstract}

%% file: sec/1_introduction.tex
\section{INTRODUCTION} \label{sec:intro}

Recent advances in CAD model-based pose estimation and tracking\cite{2022Pseudo,Random_sample_consensus,Stoiber2022IJCV,Learning_Local_RGB-to-CAD} have significantly improved object generalization capabilities, enabling robotic systems to interact with diverse objects without retraining. Of particular importance for mobile robotic applications, robust 6D pose tracking serves as the cornerstone for enabling critical manipulation tasks, including autonomous grasping and physical scene interaction on unmanned aerial vehicles (UAVs) and autonomous ground vehicles (AGVs), etc.

While existing 6D pose tracking methods demonstrate robust performance under gradual motion conditions, their accuracy deteriorates catastrophically in fast-moving scenarios characterized by rapid camera or object movements. Specifically, mainstream frameworks\cite{wen2024foundationpose,wen2020se} perform pose tracking by propagating the previous frame’s pose to predict the Region of Interest (ROI) and initialize the refinement process for the current frame. However, abrupt inter-frame movements lead to tracking failures due to two critical aspects. On the one hand, the predicted ROI, calculated from the prior pose, deviates significantly from the target’s actual location in the current frame due to sudden translational or rotational motions. Such spatial discrepancies prevent the refinement module from employing the target-related image information to compute the up-to-date pose. On the other hand, even with precise ROI localization, the refinement modules may fail to handle large inter-frame rotations due to the inherent limitations of refinement capability.

\begin{figure}[t] 
  \centering
  \includegraphics[width= 3.4in]{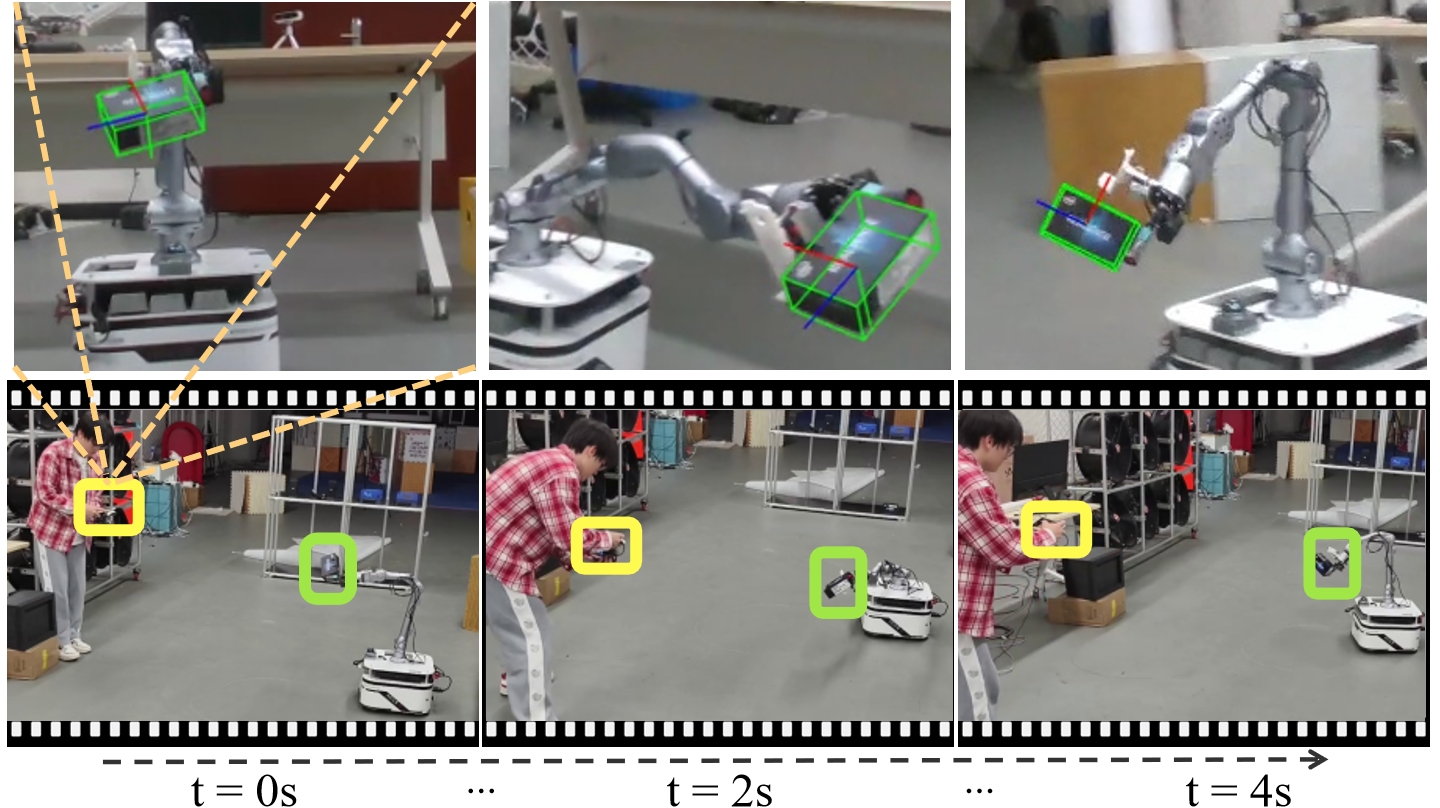}
  \caption{
  Our method, \toolName{}, achieves real-time and robust 6D object pose tracking for fast
moving camera and object.
  In real-world scenarios, a mobile manipulator controls the target object's movement, while we manually adjust the camera pose.
  This highly dynamic setup, with the object moving rapidly and camera perspective constantly changing, poses extreme challenges for 6D pose tracking. 
  The first row visualizes the object pose tracking results, where the green 3D bounding box represents the oriented bounding box of the target object, and the red, green, and blue lines indicate the XYZ axes, respectively.
  The second row shows the third-person view at the corresponding moment, with the yellow box marking the camera and the green box highlighting the target object, illustrating their rapid motion.
  Images are cropped and zoomed-in for better visualization.
  } 
  \label{top}
  \vspace{-0.5cm}
\end{figure}

To address these issues, we propose DynamicPose, a retraining-free 6D pose tracking framework that systematically improves tracking robustness through three synergistic components. Given the initial pose and CAD model of the object, this approach leverages sequential RGB-D images and onboard camera IMU data to continuously track the 6D pose of the object.
For the first issue, we introduce visual-inertial odometry (VIO), which is commonly used on mobile robots, to compensate for camera-induced ROI shifts: By estimating inter-frame camera motion through IMU-assisted visual tracking, our system maintains accurate ROI predictions even during sudden camera movements. Second, a depth-informed 2D tracker rectifies ROI deviations caused by large object translations, leveraging depth data to project 3D translations from 2D observations.
To address the second issue, we develop a VIO-guided Kalman filter for real-time object rotation tracking, enabling efficient rotation prediction during rapid motions. For extreme rotations exceeding the filter's capacity, we deploy an adaptive multi-candidate strategy that dynamically scales candidate poses based on rotational variance metrics. These candidates undergo hierarchical refinement through a pre-trained network to resolve the final 6D pose.
These components form a closed-loop system in which the 6D pose tracking results are used to guide subsequent 2D tracking and Kalman filter updates.

Extensive experiments show that \toolName{} outperforms existing SOTA methods for 6D object pose tracking in fast-moving camera and object scenarios, where the relative motion between the target object and the camera exceeds $1.5 m/s$ and $3.0 rad/s$.
As shown in Fig. \ref{top}, our method is capable of robust 6D pose tracking of the target object in real-world scenarios.
Notably, the framework maintains real-time performance without retraining.

Our contributions are summarized as follows:

\begin{itemize}
    \item \textbf{Efficient translation compensation mechanism}: can correct ROI offset caused by rapid movement of the camera and target object, ensuring the accuracy of translation initialization.
    \item \textbf{Robust rotation estimation strategy}: 
    a VIO-guided Kalman filter with dynamically scaled multi-candidate refinement, enabling robust 6D tracking under extreme rotations.
\end{itemize}
Code developed in this work will be released at \url{https://github.com/Robotics-STAR-Lab/DynamicPose}.

%% file: sec/2_related_work.tex
\section{RELATED WORK} \label{sec:related}

\subsection{CAD Model-based Object Pose Estimation}

Instance level pose estimation methods \cite{He_2020_CVPR,He_2021_CVPR,labbe2020,Park_2019_ICCV} assume a textured CAD model of 
the object is provided. These methods require training and testing to be performed on the exact same instance. The object pose can be estimated by direct regression \cite{Li_2019_ICCV,xiang2017posecnn}, or constructing 2D-3D correspondances followed by least squares fitting \cite{He_2020_CVPR,He_2021_CVPR}. Such methods are not applicable to our task, as retraining and parameter adaptation for each new object instance is impractical in real-world robotic applications, where rapid deployment and zero-shot generalization are critical requirements. Category-level methods exhibit better generalization capability compared to instance-level approaches. Such methods can be applied to novel instances of the same category, but they cannot generalize to arbitrary novel objects beyond the predefined categories. To tackle this limitation, recent works \cite{labbe2022megapose, osop, wen2024foundationpose} focus on enabling instant 6D pose estimation of arbitrary novel objects without requiring instance-specific training or parameter tuning, provided that the CAD model is available at test time. This kind of method achieves zero-shot generalization across diverse object categories while maintaining real-time tracking capabilities under dynamic conditions. These methods generate pose candidates by uniformly sampling the CAD model, retrieve the most similar candidate via template matching with the input images, and refine it through a refinement network to output the final 6D pose estimation.

\subsection{6D Object Pose Tracking}
6D object pose tracking aims to improve the accuracy and stability of pose tracking in video sequences by fusing temporal information. The current tracking method system can be classified according to the level of prior knowledge: instance-level methods (requiring known object CAD models) \cite{deng2021poserbpf,li2018deepim,wen2020se}, and category-level methods (based on object category features) \cite{lin2022icra:centerposetrack,wang20196-pack}. We focus on technical branches in instance-level tracking methods.

Some 6D tracking methods utilize feature matching to get the 6D pose of the object. This type of method constructs an objective function that reflects the discrepancy between the current observation and the previous state, and computes relative transformations based on the minima of the residual function \cite{7353536,7410443,10.1007/s11263-018-1119-x,8565885}. However, this category of 6D tracking methods fails to meet the requirements of real-time 6D pose tracking for rapidly moving objects in mobile robotic applications. Because the noise sensitivity of feature detection limits the robustness of algorithms. Besides, these algorithms require complex hyperparameter tuning for different scenarios, which makes it difficult to meet the deployment requirements of mobile robots in dynamic environments.


Other algorithms use historical pose information to estimate the pose distribution of the current frame and select the best-estimated pose based on confidence. \textit{Deng et al.} \cite{deng2021poserbpf} proposed a Rao-Blackwellized particle filter, which uses an encoder-decoder network to extract the object from the image and remove the background to better estimate the similarities between the ground-truth pose and pose templates. Although  \textit{Deng et al.} achieved excellent performance on the YCB dataset \cite{7251504,7254318}, this method is not applicable to mobile robotic scenarios because the encoder-decoder network requires retraining for every new scene. 

Recently, FoundationPose \cite{wen2024foundationpose} enables 6D pose tracking across diverse objects and scenes without requiring network retraining while maintaining state-of-the-art accuracy. FoundationPose achieves generalization of the score and refinement networks through a large amount of synthetic data training and contrast validation, enabling robust 6D pose estimation across novel objects and scenarios. However, this approach encounters significant limitations in fast-moving camera and object scenarios. Its tracking accuracy degrades in fast-moving scenarios due to two core limitations: (1) the region of interest (ROI), derived from the prior frame’s pose, drifts under rapid motion, causing misalignment with the object’s actual location and depriving the refinement module of valid visual cues; (2) the refinement network cannot handle large inter-frame rotational discrepancies, leading to refine failure. The error will accumulate and eventually cause tracking failure.
    
Therefore, to overcome these issues, we propose \toolName{}, which integrates an efficient closed-loop translation compensation mechanism to dynamically rectify the region of interest (ROI) while introducing a robust rotation estimation strategy to enhance the refinement network’s accuracy under extreme motion conditions.

%% file: sec/3_problem_formulation.tex
\section{PROBLEM FORMULATION} \label{sec:formulation}

\begin{figure*}[!t]
      \centering
       \includegraphics[scale=0.2]{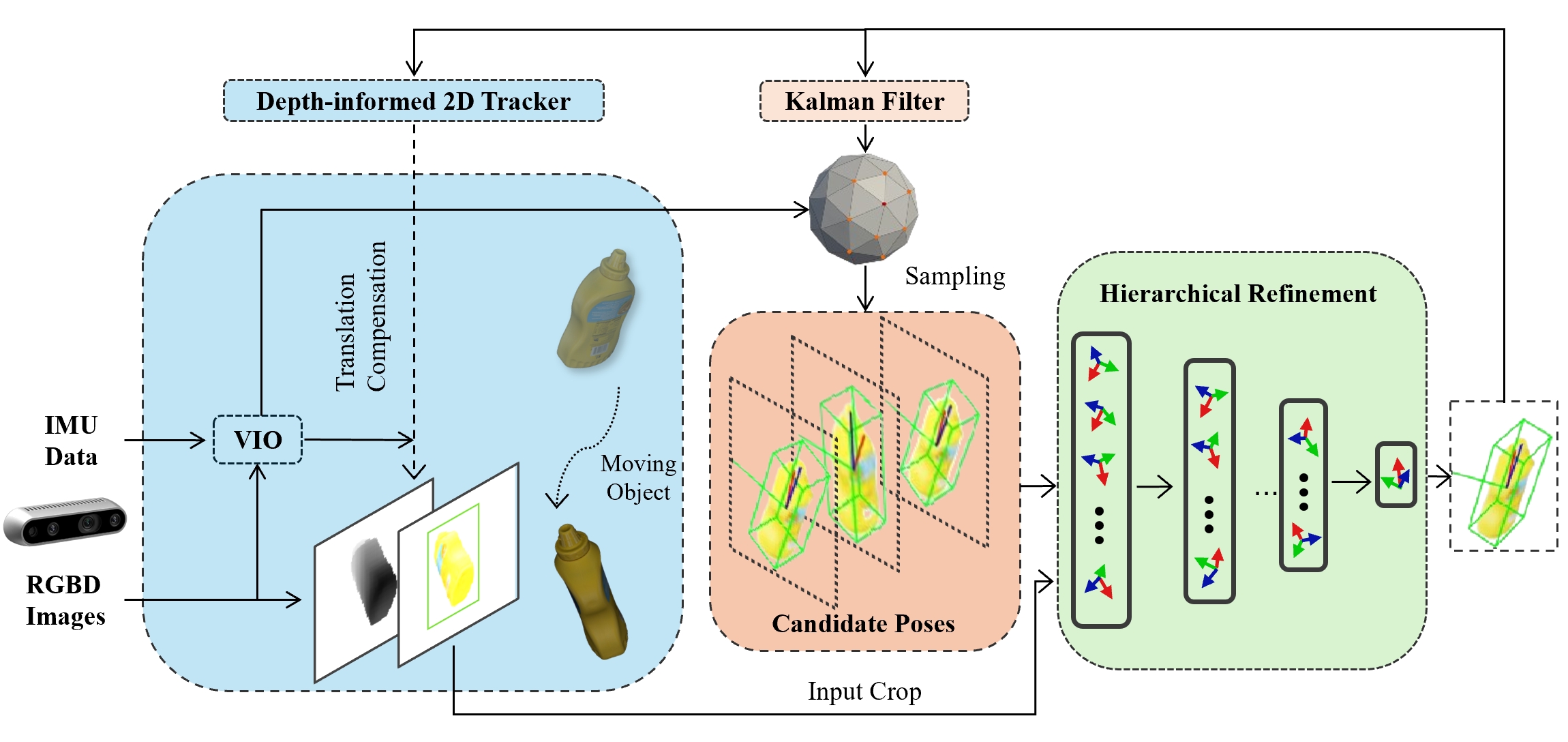}
      \caption{
  System Overview of \toolName{}.
  We introduce Visual-Inertial Odometry (VIO) to compensate for the Region of Interest (ROI) shifts and the relative rotation between the object and the camera, which are induced by camera motion.
  Then, we utilize a depth-informed 2D tracker to correct the ROI deviations resulting from large object translations. 
  A Kalman filter is employed to make coarse rotation predictions during rapid motion, and multiple candidate poses are generated based on a multi-hypothesis strategy. 
  Subsequently, a pre-trained refinement network is used for hierarchical refinement and outputs the final 6D object pose.
  The 6D pose tracking result is then fed back to the 2D tracker to guide the 2D tracking of the next frame and Kalman filter updates.
      }
      \label{overview}
      \vspace{-0.5cm}
   \end{figure*}

This study addresses the problem of CAD model-based 6D object pose tracking\cite{liu2024survey} in fast-moving camera and object scenarios. The task requires tracking the 6D pose of an object using an RGB-D data stream, using the object's CAD model and an initial pose estimation, in the presence of rapid motion of the camera and the target object. 
6D pose tracking algorithms are widely used in tasks such as object manipulation in mobile robots. Since mobile robots are typically equipped with sensors such as RGBD cameras and inertial measurement units (IMU), data such as RGB images, depth images, and relative positioning information can be utilized in our method to achieve efficient object pose tracking.

This work aims to leverage RGBD cameras and IMU to perform real-time and robust 6D pose tracking of the target object in scenarios with rapid camera and object movements, given an initial pose estimation.

For simplicity, the following symbols are defined:

We use the homogeneous transformation matrix
\[
\mathbf{T}_{B(t)}^{A(t)} = 
\begin{bmatrix}
\mathbf{R}_{B(t)}^{A(t)} & \mathbf{t}_{B(t)}^{A(t)} \\
0 & 1
\end{bmatrix}
\in SE(3)
\]
to represent the pose of coordinate frame \( B \) relative to coordinate frame \( A \) at time \( t \). 
Here, \( \mathbf{R}_{B(t)}^{A(t)} \in SO(3) \) represents the rotation, while \( \mathbf{t}_{B(t)}^{A(t)} \in R^{3} \) represents the translation.
Similarly, the homogeneous transformation matrix
\[
\mathbf{T}_{B(t)}^{A(t-1)} = 
\begin{bmatrix}
\mathbf{R}_{B(t)}^{A(t-1)} & \mathbf{t}_{B(t)}^{A(t-1)} \\
0 & 1
\end{bmatrix}
\in SE(3)
\]
is used to represent the pose of frame B at time $t$ relative to frame A at time $t-1$.

%% file: sec/4_method.tex
\section{METHODOLOGY} \label{sec:method}

\subsection{Preliminaries}
FoundationPose \cite{wen2024foundationpose} proposes a unified framework for 6D object pose estimation and tracking. In model-based settings, it initializes global poses uniformly around the object, refines them through a dedicated network, and selects the optimal pose via scoring network. 
In the object pose tracking task, FoundationPose\cite{wen2024foundationpose} initializes the pose of the current frame through the previous frame's pose and locates the Region of Interest(ROI). Then, the refinement network is used to refine the pose, obtaining the pose tracking result of the current frame.

\begin{figure}[!htb] 
  \centering
  \includegraphics[width= 3.4in]{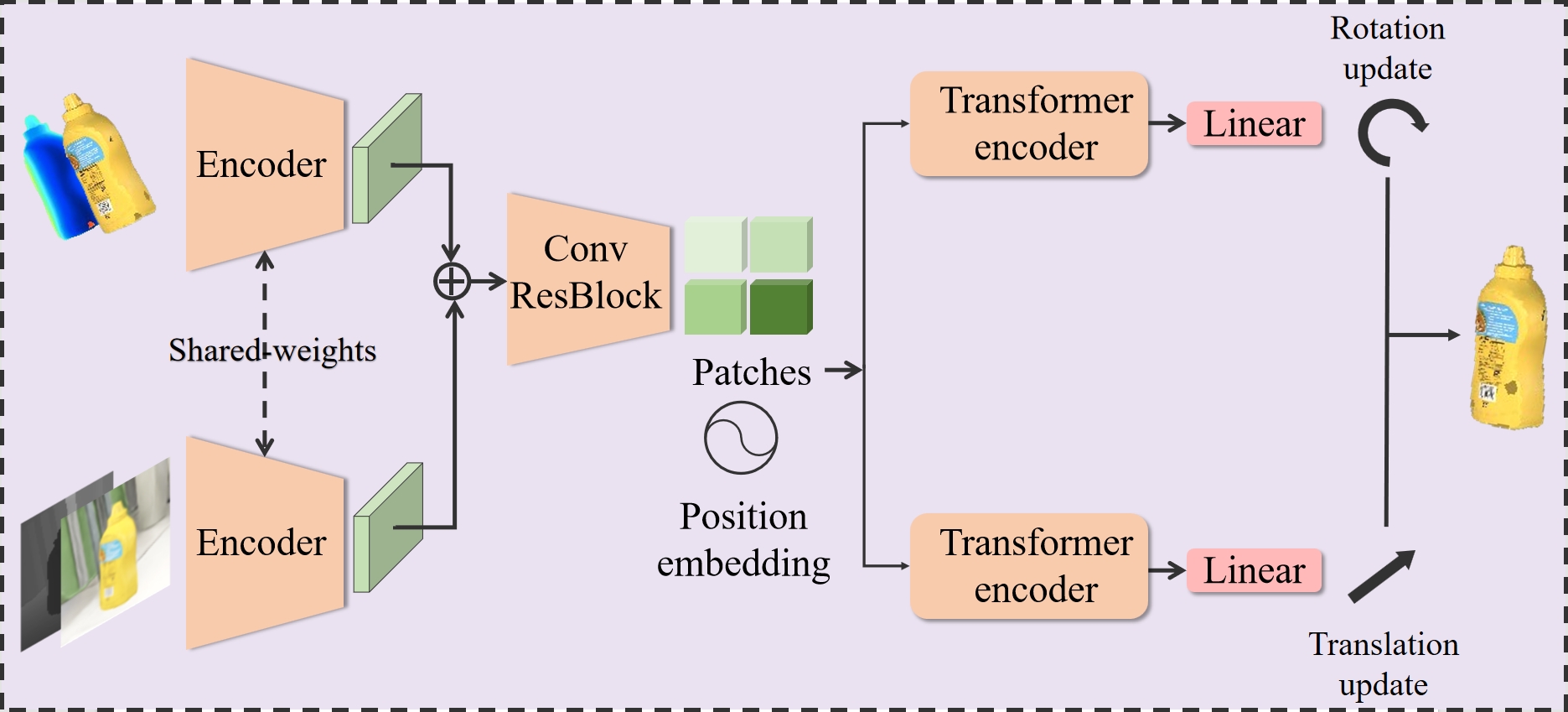}
  \caption{
  Refinement network architecture of FoundationPose\cite{wen2024foundationpose}. The refiner takes as input the rendering of the object conditioned on the coarse pose and a crop of the input observation from the camera, and outputs a pose update.
  } 
  \label{refinement}
  \vspace{-0.5cm}
\end{figure}

One of the key components of FoundationPose \cite{wen2024foundationpose} is its efficient refinement network, which can refine the coarse pose to a more precise pose. As shown in Fig.\ref{refinement}, the refinement network of FoundationPose \cite{wen2024foundationpose} takes as input the rendering of the object conditioned on the coarse pose and a crop of the input observation from the camera, extracting feature maps by a single shared CNN encoder. The feature maps are concatenated, fed into CNN blocks with residual connection \cite{Resnet}, and tokenized by dividing into patches with position embedding. Finally, after the transformer encoder and linear projection, the translation and rotation updates are predicted to obtain the refined pose. In addition, the method is trained on a large-scale synthetic dataset generated by the LLM-aided pipeline, ensuring robustness to various textures, lighting, and occlusions. This makes it perform well in pose estimation and pose tracking in static or quasi-static scenarios. Therefore, we leverage this network architecture.


However, in fast-moving camera and object scenarios involving rapid camera and object motion, FoundationPose\cite{wen2024foundationpose} exhibits limitations. Reliance on the pose of the previous frame leads to cascading failures: 
The abrupt movements cause a spatial deviation between the ROI calculated from the previous pose and the actual target location, leading to incorrect pose initialization. While the large inter-frame rotations exceed the processing capabilities of the refinement network, resulting in incorrect pose tracking results.
%
To overcome these limitations, we enhance the framework with an efficient translation compensation mechanism and a robust rotation estimation strategy. This ensures accurate region initialization and candidate pose generation, leading to reliable 6D object pose tracking.

\subsection{System Overview}

An overview of our proposed framework is shown in Fig. \ref{overview}. Specifically, we first introduce visual-inertial odometry (VIO) to compensate for the ROI offset caused by the camera motion and employ a depth-informed 2D tracker to correct the ROI deviation caused by the large object translation. Next, we utilize the relative positioning information derived from VIO to compensate for the relative rotation between the object and the camera, while employing a Kalman filter to coarsely predict the rotation. At the same time, based on the prediction results, candidate poses are generated through a multi-hypothesis strategy.
The final 6D pose tracking result is obtained through hierarchical refinement, utilizing a pre-trained refinement network.
The pose tracking result is further used to guide subsequent 2D tracking and Kalman filter updates, forming a closed-loop system.

\subsection{Pose Initialization} \label{sec:initialization}


In the pose initialization stage, given the object masks generated by SAM2\cite{ravi2024sam2}, we use FoundationPose\cite{wen2024foundationpose} to estimate the pose of the first frame to provide initial input for subsequent pose tracking. 
Specifically, based on the 2D mask and the median depth, we initialize the translation part of the pose and crop the input RGBD image.
Then, $N_s$ viewpoints are uniformly sampled on the regular icosahedron centered on the object CAD model, and $N_i$ in-plane rotations are performed on each viewpoint to generate a total of $N_s \cdot N_i$ pose templates. 
These poses are iteratively refined through the pose refiner, and the one with the highest score that best matches the input image is selected, thereby obtaining an accurate object pose estimate.

\subsection{Translation Compensation Mechanism} \label{sec:translation}
Sudden motion in fast-moving scenarios is one of the major factors for the significant degradation of 6D pose tracking performance.
In such cases, the large spatial deviation between the predicted ROI(calculated from the previous pose) and the actual target position adversely affects region initialization. 
Therefore, a deviation correction mechanism is crucial to correct the ROI offset caused by sudden motion,
ensuring a more accurate translation initialization.

\begin{figure}[t] 
  \centering
  \includegraphics[width= 3.2in]{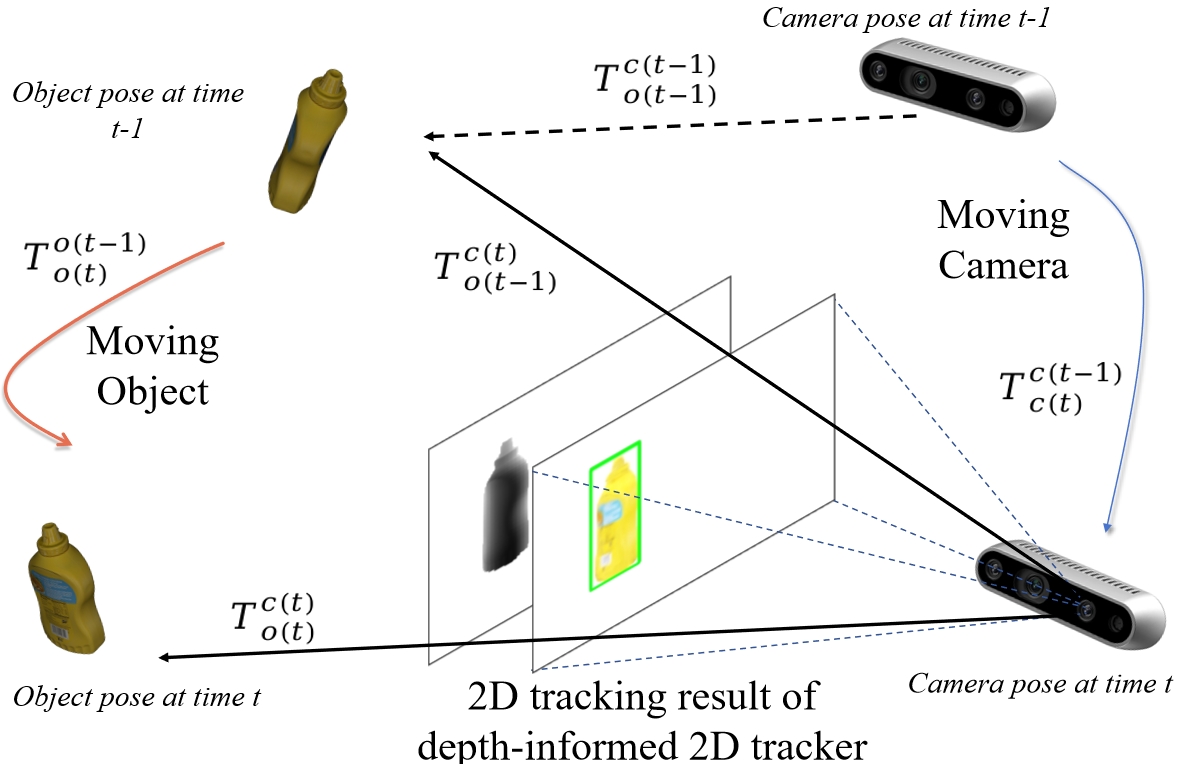}
  \caption{Diagram of translation compensation. 
   This figure illustrates the inter-frame motion between the target object and the camera under conditions of rapid movement for both the camera and the object.
  We compensate for camera-induced Region of Interest (ROI) shifts via visual-inertial odometry (VIO). 
  We rectify ROI deviations caused by large object translation by using a depth-informed 2D tracker, achieving accurate translation initialization.
  } 
  \label{translation_method}
  \vspace{-0.5cm}
\end{figure}

As shown in Fig.\ref{translation_method}, we implement an efficient translation compensation mechanism.
Specifically, we are given the pose of the target object in the camera coordinate system at the previous frame, $\mathbf{T}_{o(t-1)}^{c(t-1)} $, which represents the object's pose at time $t-1$ relative to the camera at time $t-1$.
Through VIO\cite{qin2019b}, we can easily obtain the transformation $\mathbf{T}_{c(t-1)}^{c(t)}$, which describes the camera's pose at time $t-1$ relative to the camera at time $t$. 
We can then compute:
\begin{equation}
    \mathbf{t}_{o(t-1)}^{c(t)} = \mathbf{R}_{o(t-1)}^{c(t-1)} \cdot \mathbf{t}_{c(t-1)}^{c(t)} + \mathbf{t}_{o(t-1)}^{c(t-1)}
    \label{eq:camMotion}
\end{equation} where $\mathbf{t}_{o(t-1)}^{c(t)}$ means the object translation at time $t-1$ relative to the camera at time $t$.
When the translation of the object is negligible, i.e., $\mathbf{t}_{o(t)}^{o(t-1)} = \mathbf{0}$, where $\mathbf{t}_{o(t)}^{o(t-1)}$ is the object's translation at time $t$ relative to the object at time $t-1$.
We can get the corrected translation $\mathbf{t}_{o(t)}^{c(t)}$ by:
\begin{equation}
    \mathbf{t}_{o(t)}^{c(t)} = \mathbf{R}_{o(t-1)}^{c(t)} \cdot \mathbf{t}_{o(t)}^{o(t-1)} + \mathbf{t}_{o(t-1)}^{c(t)} = \mathbf{t}_{o(t-1)}^{c(t)}
    \label{eq:vioForTranlation}
\end{equation}
where $\mathbf{t}_{o(t)}^{c(t)}$ is the object's translation at time $t$ relative to the camera at time $t$.This calculation enables us to compensate for the ROI shift caused by the camera motion.
However, the above process may encounter errors when handling large object translation, where $\mathbf{t}_{o(t)}^{o(t-1)} \neq \mathbf{0}$.
We employ a depth-informed 2D tracker to solve ROI misalignment caused by large object translation. Based on the target object's location in the previous RGB frame, the 2D tracker tracks the 2D bounding box of the object in the current frame. Leveraging the centroid depth value, the 2D observation is projected into 3D camera coordinates by: 
\begin{equation}
\mathbf{t}_{o(t)}^{c(t)} =\mathbf{K}^{-1} \cdot [u, v, 1]^{T} \cdot z
\label{eq:2dForTranlation}
\end{equation}
where $\mathbf{K} \in R^{3 \times 3}$ is the camera intrinsic matrix, $(u, v)$ are the center coordinates of the 2D bounding box in the RGB image, and $z$ is the corresponding median depth value.

Equation \ref{eq:vioForTranlation} and \ref{eq:2dForTranlation} enable us to correct the ROI deviation and accurately initialize the object's translation in the camera coordinate system, whether both the object and camera are moving rapidly or only the camera is moving rapidly.

\subsection{Rotation Estimation Strategy} \label{sec:rotation}

\begin{figure}[t] 

  \centering
  \includegraphics[width=3.4in]{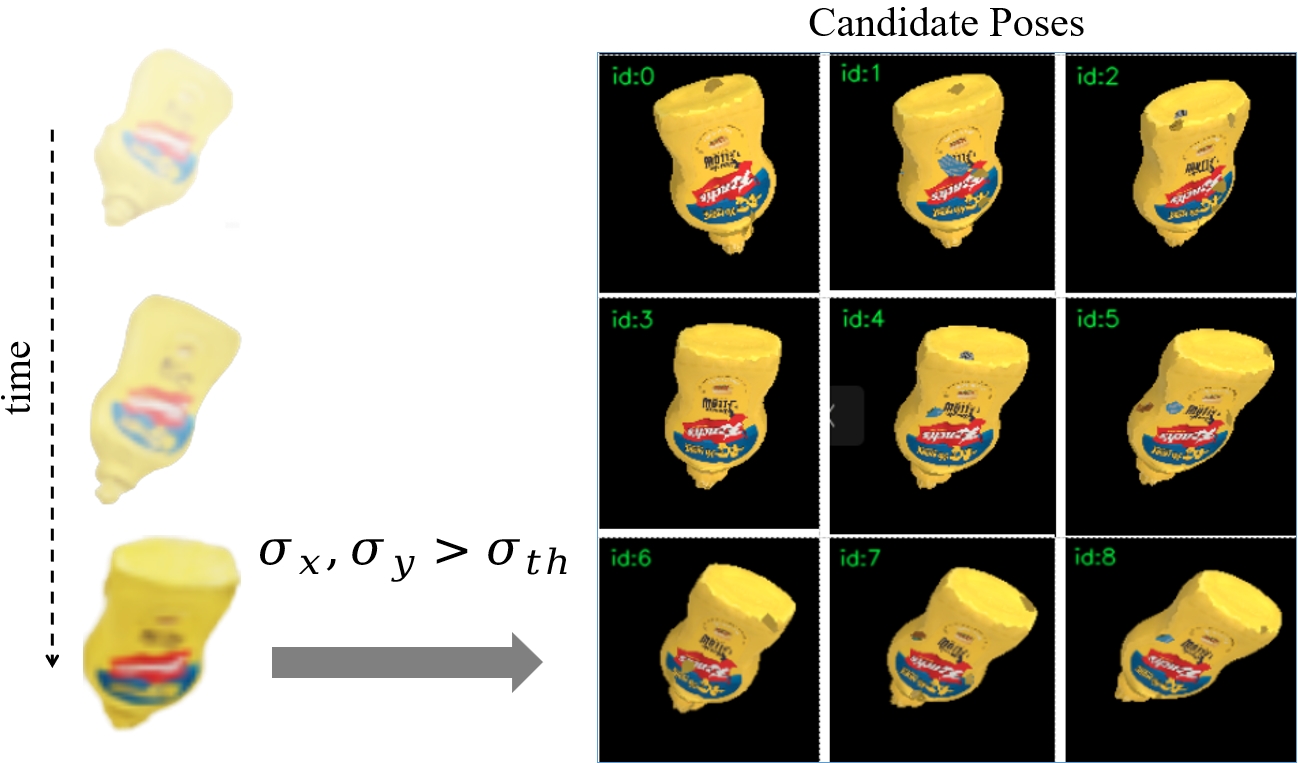}
  \caption{Diagram of rotation estimation. Leveraging historical pose information, we use the Kalman filter to predict the rotation for the current frame, and dynamically adjust the generation of candidate poses based on the variance of the rotation distribution, providing reliable rotation estimation.
  } \label{rotation_methoddd}
  \vspace{-0.5cm}
\end{figure}

In fast-moving scenarios where both robots and objects undergo extensive and abrupt motion, employing the pose candidate derived solely from the estimation of the previous frame or a constant velocity model prediction is inadequate. These approaches result in substantial rotation discrepancies between the pose candidate and the ground-truth object pose, which severely compromises the refinement network's capacity to achieve accurate pose estimation. To tackle this issue, we propose a VIO-guided Kalman filter framework for generating candidate poses. This framework compensates the ego-motion of the camera, dynamically adjusts the number of candidate poses according to the Kalman filter's prediction covariance, and selects proper candidates for further refinement.

\textbf{Ego-motion Compensation.} 
In fast-moving scenarios where both the camera and target object undergo large motion, predicting the object's motion state within a moving camera frame becomes inherently challenging and inaccurate. This instability degrades the Kalman filter's ability to predict rotational states of moving objects. To address this limitation, we propose anchoring target objects within the world coordinate by integrating visual-inertial odometry (VIO). This can stabilize the rotational estimation process under aggressive motion conditions and enable robust 6D pose tracking.

\begin{equation}
   \mathbf{R}_{o(t)}^{w} = \mathbf{R}_{c(t)}^{w} \cdot \mathbf{R}_{o(t)}^{c(t)},  \label{sec:VIO}
\end{equation}
where $\mathbf{R}_{o(t)}^{c(t)}$ is the object's rotation in the camera frame at time $t$. 

\textbf{Angular Velocity Prediction.} 
The object's motion between consecutive frames can be approximated as constant velocity due to the minimal inter-frame time difference. We use Tait–Bryan angles $\mathbf{\theta}=[\theta_{roll}, \theta_{pitch}, \theta_{yaw}]$ to represent rotation, then we use a constant angular velocity model to predict Tait–Bryan angles at the next time step:
\begin{equation}
    \mathbf{\omega}(t)=\frac{\mathbf{\theta}(t)-\mathbf{\theta}(t-1)}{\Delta t}
\end{equation}
\begin{equation}
    \mathbf{\theta}(t+1) = \mathbf{\theta}(t) + \mathbf{\omega}(t)\cdot\Delta t
\end{equation}
where $\mathbf{\omega}=(\omega_x, \omega_y, \omega_z)$ represents angular velocity.

However, the constant angular velocity model is inaccurate when the object is undergoing inconstant angular velocity. The predicted pose using a constant angular velocity model may differ significantly from the true values, resulting in the refinement network's inability to accurately estimate the current 6D pose of the object. To address this issue, we introduce a Kalman filter to describe the change in the object's pose. 

\textbf{Kalman Filter Setups.} The Kalman filter tracks only the object's orientation in Tait–Bryan angles $\mathbf{\theta} = [\theta_\text{roll}, \theta_\text{pitch}, \theta_\text{yaw}]$. We assume that the object's XYZ Tait–Bryan angle $\theta_\text{roll}$, $\theta_\text{pitch}$, and $\theta_\text{yaw}$ follow independent normal distributions with means $\mu_{\text{roll}}$, $\mu_{\text{pitch}}$, $\mu_{\text{yaw}}$ and variances $\sigma_{\text{roll}}^2$, $\sigma_{\text{pitch}}^2$, $\sigma_{\text{yaw}}^2$, respectively. 

The predicted Tait–Bryan angles at $t+1$ are:
\begin{equation}
    \mathbf{\theta}^-(t+1)=\mathbf{\theta}^-(t)+\mathbf{\omega}(t)\Delta t +\mathbf{q_{\theta}},
\end{equation}
where $\mathbf{\theta^-}, \mathbf{\theta^+}$ represent the predicted pose and the updated pose, respectively. $q_{\theta}\sim \mathcal{N}(0, \mathbf{Q})$ is process noise with covariance:
\begin{equation}
    \mathbf{Q}=\|{\dot{\mathbf{\omega}}(t)}\| \cdot \mathbf{I},
\end{equation}
and ${\dot{\mathbf{\omega}}(t)}=\frac{\mathbf{\omega}(t)-\mathbf{\omega}(t-1)}{\Delta t}$ is the angular acceleration. The noise covariance scales with acceleration magnitude to adapt to inconstant angular velocity.

As for measurement update, given an observed Tait–Bryan angle $\mathbf{\theta_\text{obs}}$, which is refined by the refinement network, the Kalman filter updates the state and covariance $\mathbf{P}$ via:
\begin{equation}
    \mathbf{K}=\mathbf{P^-H^\top}(\mathbf{HP^-H^\top+\mathbf{R})}^{-1},
\end{equation}
\begin{equation}
    \mathbf{\theta^+} = \mathbf{\theta^-}+\mathbf{K}(\mathbf{\theta}_\text{obs}-\mathbf{H\theta^-}) ,
\end{equation}
where H is a $3\times3$ identity matrix, $\mathbf{R}$ is measurement noise covariance, and $\mathbf{K}$ is the Kalman gain. Based on the assumption of independent rotations about the three axes, the measurement noise covariance matrix R  constitutes a diagonal matrix. Under conditions of robust tracking where measurements exhibit high accuracy, the measurement noise variance can be assigned a minimal value. For our experiments, this parameter is set to 0.05.

\textbf{Adaptive Candidate Poses Sampling.} To balance computational efficiency and tracking robustness under varying motions, we propose a variance-aware sampling strategy that dynamically adjusts the number of candidate poses based on motion uncertainty. This adaptation allows the algorithm to make more samples along axes that exhibit higher motion variability (e.g. sudden rotational changes) while applying fewer samples to axes with stable motion patterns. We define a threshold $\sigma_{th}$ to measure motion stability. For each rotational axis $i \in \text{\{roll,pitch,yaw\}}$, the number of sampled points $N_i$ along axis $i$ is determined by:
\begin{equation}
N_i = 
    \begin{cases}
1, & \text{if } \sigma_i < \sigma_{th} \text{ (constant angular velocity motion)} \\
3, & \text{if } \sigma_i \geq \sigma_{th} \text{ (inconstant angular velocity motion)}
\end{cases}
\end{equation}
This can reduce samples when motion is close to constant velocity motion, thereby improving the efficiency of the algorithm.

For each axis $i$:
\begin{equation}
\theta_{i,cand}=
    \begin{cases}
    \mu_i, & \text{if }\sigma_i<\sigma_{th}\\
    {\mu_i-\sigma_i,\mu,\mu+\sigma_i}, &\text{if } \sigma_i \geq \sigma_{th}
    \end{cases}
\end{equation}
where $\mu$ refers to the mean value. candidate poses are generated through systematic combinations of sampled points along each rotational axis (roll, pitch, yaw).

\subsection{Hierarchical Refinement} \label{sec:refine}
The candidate poses generated in the previous step are affected by noise, so it is necessary to improve the pose quality further and select the best pose output. 
To this end, we adopt a hierarchical refinement, combined with a pre-trained refinement network, to iteratively improve the candidate poses and obtain the best result through layer-by-layer screening.
Specifically, we use the refiner module in FoundationPose\cite{wen2024foundationpose}, which takes as input the rendering image of the object on the coarse pose and the current camera observation. The refiner outputs a pose update to improve the pose quality.
In the iteration $i$ (where $0 <= i <= N$, $N$ is the maximum number of iterations), the output of the network includes the translation update $\Delta \mathbf{t}_{i}$ and the rotation update $\Delta \mathbf{R}_{i}$. The input $K_i$ candidate poses 
$
\mathbf{[R_i | t_i]} \in SE(3)
$ 
are improved by:
\begin{equation}
\mathbf{t_{i+1}} = \mathbf{t_{i}} + \mathbf{\Delta t_{i}}
\end{equation}
\begin{equation}
\mathbf{R_{i+1}} = \mathbf{\Delta R_{i}} \otimes \mathbf{R_{i}}
\end{equation}
Here,  $\otimes$ denotes update on $SO(3)$, $\mathbf{t_{i+1}}$ and $\mathbf{t_{i}}$  represent the translation for the next and current frames, respectively, while $\mathbf{R_{i+1}}$ and $\mathbf{\Delta R_{i}}$ represent the rotation for the next and current frames, respectively.

To measure the closeness between the current predicted pose and the observed true pose, we use the $L2$ norm $\| \Delta \mathbf{R_{i}} \|_2$ of the rotation update output by the refiner as the metric. 
A smaller value indicates that the predicted pose is closer to the true pose. 
This is because when the difference between the predicted and true poses is larger, a greater rotation update is typically required. 
In contrast, when they are closer, the rotation update becomes smaller.
After completion of this refinement, we select the top $K_{i+1}$ candidate poses with the smallest rotation update ($K_{i+1}$ is the number of candidate poses for the next round of iteration, and $K_{N}=1$) to enter the next round of improvement.
Through the hierarchical refinement process, the candidate poses generated in Fig. \ref{rotation_methoddd} are further refined, and the pose closest to the true pose is selected layer by layer as the final 6D object pose tracking output.

Finally, we calculate the projected region of the 6D pose tracking result in the RGB image and use it to update the target template of the depth-informed 2D tracker. This helps guide the tracking of the target object in the next frame, providing a more accurate 2D bounding box. At the same time, we use the pose result to update the Kalman filter, aiding in the accurate rotation prediction for the next frame.

%% file: sec/5_experiments.tex
\section{EXPERIMENT} \label{sec:exp}

\subsection{Setup}


In the experimental section, we will present the performance of our 6D object pose tracking framework, \toolName{}, in simulation environments. Since existing datasets do not provide camera poses or IMU data, we are unable to compare with baseline algorithms on them.
To evaluate the tracking capability of our approach for different objects, we select 13 different objects from the YCB-Video dataset and perform tests in the Isaac Sim simulation environment. In the simulation, both objects and the camera are set to move rapidly along specified trajectories, where the relative velocity exceeds $1.5m/s$, and the angular velocity exceeds $3.0rad/s$ between the target object and the camera.

\begin{table}[t]
\setlength{\tabcolsep}{1.5pt} 
\renewcommand{\arraystretch}{1.5}
\centering
\caption{Pose Tracking Results in Simulation Environment}
\label{baselines_results}
\begin{tabular}{l|cc|cc|cc}
\hline\hline
 & \multicolumn{2}{c|}{\textbf{FoundationPose\cite{wen2024foundationpose}}}
 & \multicolumn{2}{c|}{\textbf{SRT3D\cite{Stoiber2022IJCV}}} 
 & \multicolumn{2}{c}{\textbf{Ours}} \\
\hline
\textbf{Metrics} & ADD-S & ADD & ADD-S & ADD & ADD-S & ADD\\ 
\hline
002\_master\_chef\_can & 13.0 & 12.1 & 53.5 & 9.4 & \cellcolor{red!20} 89.4 & \cellcolor{red!20} 76.1  \\
003\_cracker\_box & 42.2 & 40.0 & 11.9 & 6.3 & \cellcolor{red!20} 89.7 & \cellcolor{red!20}74.0  \\
005\_tomato\_soup\_can & 12.8 & 11.9 & 1.9 & 1.6 & \cellcolor{red!20} 88.9 & \cellcolor{red!20} 77.0  \\
006\_mustard\_bottle & 13.4 & 12.7 & 29.5 & 7.6 & \cellcolor{red!20}92.4 & \cellcolor{red!20}84.2  \\
009\_gelatin\_box & 13.5 & 12.7 & 3.2 & 2.7 & \cellcolor{red!20}88.6 & \cellcolor{red!20}76.9  \\
010\_potted\_meat\_can & 13.3 & 12.4 & 8.0 & 4.8 & \cellcolor{red!20}90.0 & \cellcolor{red!20}78.3  \\
011\_banana & 13.1 & 12.2 & 3.3 & 2.8 & \cellcolor{red!20}87.5 & \cellcolor{red!20}75.3  \\
019\_pitcher\_base & 23.1 & 21.8 & 7.6 & 5.6 & \cellcolor{red!20}86.5 & \cellcolor{red!20}69.0  \\
025\_mug & 13.7 & 12.7 & 42.0 & 12.4 & \cellcolor{red!20}90.0 & \cellcolor{red!20}76.5 \\
035\_power\_drill & 13.7 & 12.8 & 23.5 & 15.4 & \cellcolor{red!20}87.3 & \cellcolor{red!20}74.3 \\
036\_wood\_block & 23.8 & 22.2 & 46.7 & 7.8 & \cellcolor{red!20}88.6 & \cellcolor{red!20}75.3  \\
052\_extra\_large\_clamp & 13.4 & 12.5 & 2.7 & 2.2 & \cellcolor{red!20}87.1 & \cellcolor{red!20}70.0 \\
061\_foam\_brick & 13.0 & 12.1 & 22.1 & 9.0 & \cellcolor{red!20}90.1 & \cellcolor{red!20}78.2 \\
\hline
MEAN & 17.1 & 16.0 & 19.7 & 6.7 & \cellcolor{red!20} \textbf{88.9} & \cellcolor{red!20} \textbf{75.8} \\
\hline\hline
\end{tabular}
\end{table}

\subsection{Metrics}

We use the average distance (ADD) metric \cite{10.1007/978-3-642-37331-2_42} for evaluation. 
Given the ground truth rotation $\mathbf{R}_\text{gt}(t_k)$ and translation $\mathbf{t}_\text{gt}(t_k)$ and the tracking rotation $\mathbf{R}(t_k)$ and translation $\mathbf{t}(t_k)$, the average distance computes the mean of the pairwise distances between the 3D model points transformed according to the ground truth pose and the tracking pose:
\begin{equation}
    \text{ADD}=\frac{1}{m}\sum_{x \in M}\|(\mathbf{R}_\text{gt}(t_k)\mathbf{x}+\mathbf{t}_\text{gt}(t_k))-(\mathbf{R}(t_k)\mathbf{x}+\mathbf{t}(t_k))\|
\end{equation}
where $M$ denotes the set of 3D model points and m is the number of points. For symmetric objects like bowl and foam brick, the matching between points is ambiguous for some views. Therefore, the average distance is computed using the closest point distance:

\begin{align}
    \text{ADD-S} = \frac{1}{m}\sum_{\mathbf{x}_1 \in M} \min_{\mathbf{x}_2 \in M} 
    \biggl\| & \left( \mathbf{R}_\text{gt}(t_k)\mathbf{x}_1 + \mathbf{t}_\text{gt}(t_k) \right) \nonumber \\
             & - \left( \mathbf{R}(t_k)\mathbf{x}_2 + \mathbf{t}(t_k) \right) \biggr\|
    \label{eq:add-s}
\end{align}


We consider the area under the curve (AUC) of ADD and ADD-S, where we vary the threshold for the average distance. And the maximum threshold is set to $0.1m$, as suggested in \cite{xiang2017posecnn}.

\subsection{Pose Tracking Comparison}

\begin{figure}[t] 
  \centering
  \includegraphics[width= 3.4in]{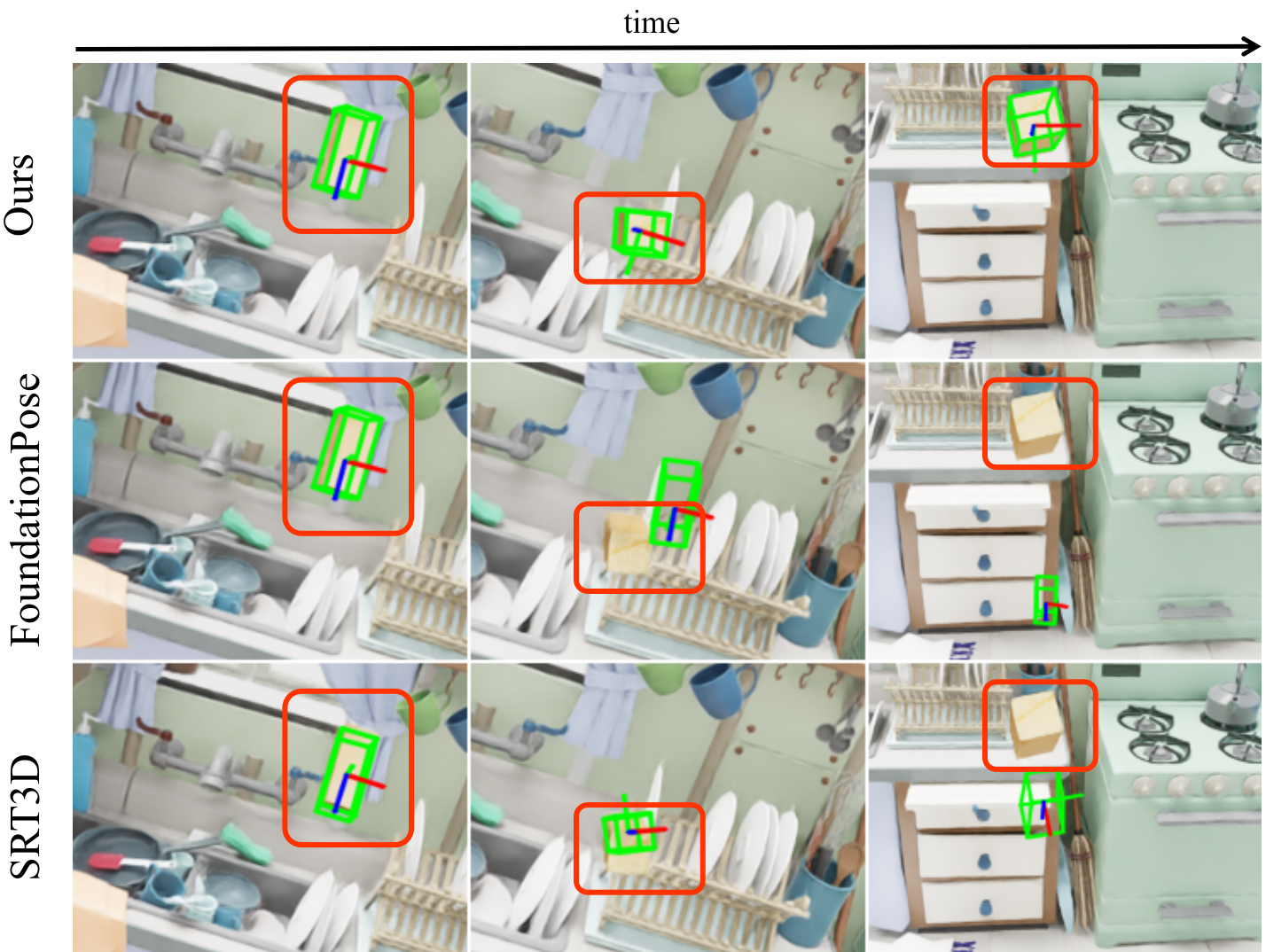}
  \caption{
  Comparison of different methods. Using object \textit{036\_wood\_block} as an example, we conduct experiments with various methods under the same rapid relative motion between the target object and the camera (from left to right represents the time sequence). Our method(first row in the figure) achieves robust 6D pose tracking. FoundationPose\cite{wen2024foundationpose}(second row) fails to track correctly due to invalid region initialization during the process. 
  SRT3D\cite{Stoiber2022IJCV}(third row) loses tracking caused by extreme inter-frame rotations.
  The red box indicates the position of the target object in the RGB image.
  Images are cropped and zoomed-in for better visualization.
  }
  \label{baseline_results_fig}
  \vspace{-0.5cm}
\end{figure}


In our experiments, we selected FoundationPose\cite{wen2024foundationpose}, which achieves the best performance among many 6D pose tracking algorithms for novel objects, and SRT3D\cite{Stoiber2022IJCV}, which is claimed to be robust to motion blur and suitable for tracking fast-moving objects, as baseline methods. 
For different objects, each method is experimented in the same scenarios, ensuring a fair comparison. 
TABLE \ref{baselines_results} presents the performance metrics of our method compared to the two baseline approaches across different objects.
Fig. \ref{baseline_results_fig} illustrates the runtime performance of the three methods on the \textit{036\_wood\_block} object.
The results show that, in the fast-moving scenarios, our method significantly outperforms the two baseline methods. 

Furthermore, as demonstrated in Fig. \ref{top}, we validate our method in real-world scenarios to assess its real-time capability and robustness for 6D object pose tracking. In these experiments, the target object is rigidly mounted on the gripper of a mobile manipulator, which is controlled to follow given trajectories with combined large translational and rotational motions. The camera pose is adjusted manually by an experimenter to simulate aggressive camera motion while ensuring the target object remains within the camera’s field of view. Experimental results demonstrate that our algorithm maintains robust and accurate pose tracking performance in fast-moving real-world scenarios.

\subsection{Analysis}
\begin{figure}[t] 
\label{fig:ablation}
  \centering
      \includegraphics[width= 3.4in]{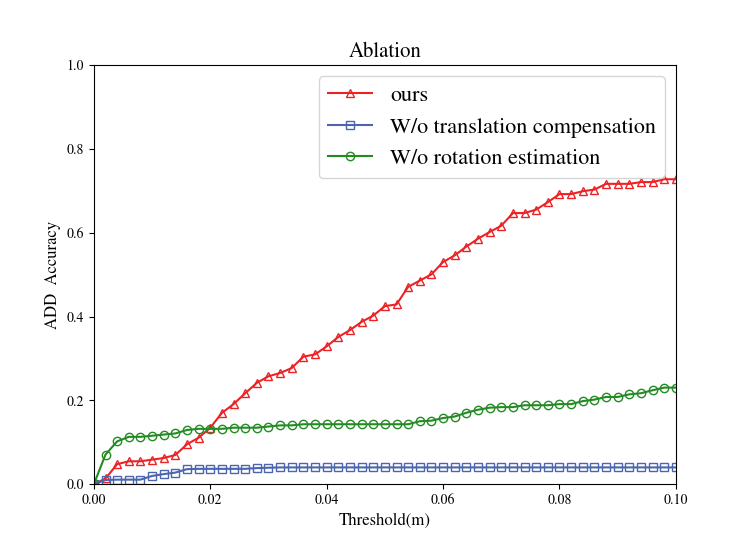}
  \caption{
  Results of Ablation Studies. 
  We show the area under the curve(AUC) of the ADD metric, where the maximum threshold is set to 0.1m.
  } 
  \label{Ablation_results}
  \vspace{-0.5cm}
\end{figure}

\textbf{Ablation Studies.} 
To evaluate the performance limits of our method and the individual contributions of components, we constructed a more dynamic simulation scenario with more aggressive motion between the target object and the camera, where the relative velocity exceeds $3m/s$ and the relative angular velocity is over $8rad/s$. 
The Fig. \ref{Ablation_results} presents our ablation study, comprising three configurations:
1) \textit{W/o translation compensation} disables the depth-informed 2D tracker;
2) \textit{W/o rotation estimation} removes the VIO-guided Kalman filter for generating candidate poses;
3) \textit{Ours} means the complete system.
The results are evaluated by the AUC of ADD metric.

\textbf{Running Time.}
We measure the running time on the hardware of Intel i7-13700F CPU and NVIDIA RTX 3060 GPU. Our approach runs at $\sim$10Hz on average.

\textbf{Limitations.}
Aggressive camera and object motions induce motion blur in RGB images, which degrades surface texture details and consequently reduces pose refinement accuracy. Besides, when handling certain objects, relying on the median depth as a translation hypothesis may introduce notable deviations from the actual translation, which can affect the accuracy of the compensation. Also, due to the limited effective range of depth cameras, inaccurate depth measurements occur for objects beyond 3 meters, significantly compromising translational estimation precision at long distances.

%% file: sec/6_discussion_and_future_work.tex
\section{CONCLUSIONS} \label{sec:discuss}
In this work, we present a novel framework for robust 6D pose tracking on mobile robotic platforms, capable of maintaining continuous and accurate pose estimation in fast-moving camera and object scenarios. The results of the experiment demonstrate that our algorithm achieves strong generalization in diverse objects without retraining while significantly outperforming the baseline methods in both tracking stability and precision. In future work, we plan to integrate our framework with robotic manipulation systems to enable dexterous object interaction in unstructured dynamic environments, further bridging the gap between pose estimation and real-world robotic applications.



